%% file: paper.tex
\documentclass{article}
\pdfoutput=1

\usepackage{graphicx,amsmath,amssymb} 
\usepackage{subfigure} 

\usepackage{algorithm}
\usepackage{algorithmic}

\usepackage{hyperref}

\long\def\ignore#1{}

\begin{document} 

\newcommand{\vek}[1]{{\bf {#1}}}
\newcommand{\lab}{{L}}
\newcommand{\labelspace}{{\cal L}}
\newcommand{\unl}{{U}}
\newcommand{\aset}{{\cal A}}
\newcommand{\cset}{{\cal C}}
\newcommand{\vx}{{\vek{x}}}
\newcommand{\vy}{{\vek{y}}}
\newcommand{\vY}{{\vek{Y}}}
\newcommand{\YA}{{\cal Y_A}}
\newcommand{\YSp}{{\cal Y}}
\newcommand{\vX}{{\vek{X}}}
\newcommand{\vr}{{r}}
\newcommand{\vv}{{\vek{v}}}
\newcommand{\vz}{{\vek{z}}}
\newcommand{\vtheta}{{\vek{\theta}}}
\newcommand{\thetaC}{{\theta^{(s,i)}}}
\newcommand{\thetaCc}{{\theta^{(s,i)}_c}}
\newcommand{\vthetaC}{{\vek{\theta^{(s,i)}}}}
\newcommand{\vc}{{\vek{c}}}
\newcommand{\vw}{{\vek{w}}}
\newcommand{\vW}{{\vek{W}}}
\newcommand{\vF}{{\vek{f}}}
\newcommand{\LL}{{\text{LL}}}
\newcommand{\map}{{\pi}}
\newcommand{\prob}{{\text{P}}}
\newcommand{\marg}{{\mu}}
\newcommand{\cliqueA}{{Clique}}
\newcommand{\nodeA}{{Node}}
\newcommand{\pairA}{{Instance Pair}}
\newcommand{\cliqueAS}{{Clique}}
\newcommand{\nodeAS}{{Node}}
\newcommand{\pairAS}{{Pair}}
\newcommand{\fullBP}{{Full}}
\newcommand{\trwOne}{{TR1}}
\newcommand{\dist}{{PR}}
\newcommand{\jordan}{{TR1}}
\newcommand{\FewM}{{Few}}
\newcommand{\ManyM}{{C}}

\title{Joint Structured Models for Extraction from Overlapping Sources}

\author{Rahul Gupta\\ IIT Bombay, India\\ \texttt{grahul@cse.iitb.ac.in} \and
Sunita Sarawagi\\ IIT Bombay, India\\ \texttt{sunita@cse.iitb.ac.in}}


\date{}
\maketitle
\input{abstract}
\section{Introduction}
\input{intro}

\section{Collective training}
\begin{figure*}
\begin{center}
\includegraphics[width=\textwidth]{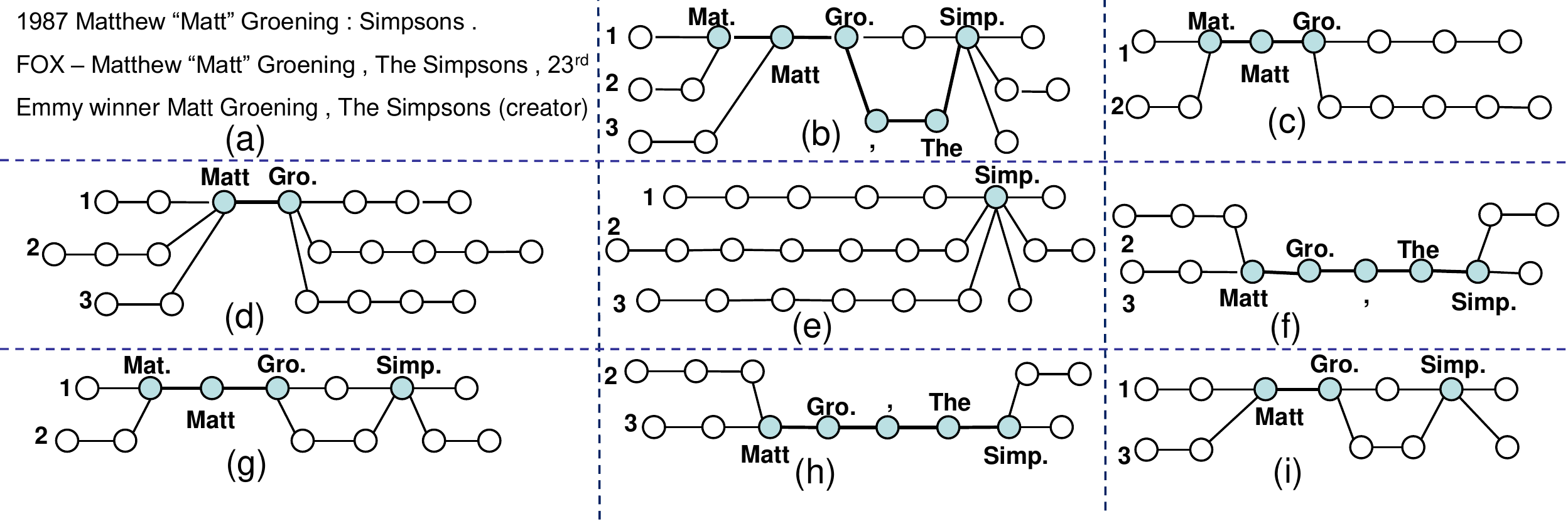}
\caption{(a) Three samples sentences with $\aset$=\{(Matthew Matt Groening), (Matt Groening), (Matt Groening , The Simpsons), (Simpsons)\} 
(b) The fused graph (c)-(f) Clique Agreement approximation (g)-(i) Instance Pair approximation.}
\label{fig-example}
\end{center}
\vskip -0.2in
\end{figure*}

Our goal is to collectively train $S$ structured prediction models,
where every source $s\in S$ comes with a small set $\lab_s$ of labeled
instances and many unlabeled instances $\unl_s$.  The unlabeled
instances from different sources share overlapping content and we seek
to exploit this for better training. For extraction tasks, this
overlap can be anywhere from the level of unigrams to non-contiguous
segments, as illustrated in Figure~\ref{fig-example}(a).  An overlap
finding algorithm identifies such shared parts as an agreement set
$\aset$ comprising of a set of cliques. Each clique $C \in \aset$
contains a list of triples $(s,i,\vr)$ indicating the source $s$, the
instance $i\in s$, and the part $\vr\in i$ which has the same content
as the other members of the clique. We do not make any assumption of
mutual exclusion between cliques and in general each clique can span a
variable number of members and token positions.  In
Section~\ref{sec-agree} we present our strategy for computing the
agreement set.

As in standard structured learning, we define a probabilistic model
$\prob_s(\vy|\vx)$ for each source $s$ using its feature vector
$\vF_s(\vx,\vy)$ and parameters $\vw_s$ as:
\begin{equation}
\label{crf-eq}
\prob_s(\vy|\vx,\vw_s)=\frac{1}{Z(\vx,\vw_s)}\exp(\vw_s\cdot\vF_s(\vx,\vy))  
\end{equation}

The traditional goal of training is to find a $\vw_s$ that 
maximizes the regularized likelihood of the labeled set in source $s$:
\begin{equation}
\LL_s(\lab_s,\vw_s) = \sum_{(\vx,\vy) \in \lab_s} \log\prob_s(\vy|\vx,\vw_s) - \gamma R(\vw_s)
\end{equation}

We propose to augment this base objective with the likelihood that the
$S$ models agree on the labels of cliques in the agreement set
$\aset$.
We first observe that the joint distribution over the labels $\vY$ of all
instances $\vX$ spanning all sources is:
\begin{equation}
\label{eq-joint}
\prob(\vY|\vX,\vW) \triangleq \prod_{s\in S}\prod_{i\in s}\prob(\vY_{si}|\vX_{si},\vw_s)
\end{equation}
where $\vW$ denotes $(\vw_1,\ldots \vw_S)$, $\vX_{si}$ represents
instance $i$ in source $s$, and $\vY_{si}$ is a random variable for
the structured output of this instance. 

Now given an agreement set $\aset$, consider the subset $\YA$ of all
possible labelings that are consistent with it:
\begin{equation}
  \YA \triangleq \{\vY : \forall C \in \aset, (s,i,\vr), (s',i',\vr') \in C: \vY_{si\vr}  = \vY_{s'i'\vr'} \} 
\end{equation}
The log likelihood of the agreement set is then:
\begin{equation}
\label{eq-agree}
\LL(\YA,\vW) \triangleq \log\Pr(\YA) =\log \sum_{\vY\in\YA}\prod_{s,i\in s}\prob(\vY_{si}|\vX_{si},\vw_s)
\end{equation}
In the rest of the paper we use the short form
$(s,i)$ to denote the instance $i$ in source $s$.
Our goal now is to jointly train $\vw_1,\ldots,\vw_S$ so as to
maximize a weighted combination of the likelihoods of the labeled and agreement sets:
\begin{equation}
\label{eq-collective}
\ignore{\textrm{\text{(P1)}}}\max_{\vw_1,\ldots,\vw_S} \sum_s \LL(\lab_s,\vw_s) + \lambda \LL(\YA,\vW)
\end{equation}

\subsection{Computing $\LL(\YA,\vW)$}
\ignore{
For agreement sets containing arbitrarily overlapping cliques spanning
many sources, the computation of $\LL(\YA,\vW)$ via
Equation~\ref{eq-agree} is not tractable.  
 computation of $\LL(\YA,\vW)$ can be
intractable.  We show that computing $\LL(\YA,\vW)$ is equivalent to
computing the log partition function in a
  However, we show next that it is possible to resort to
approximations developed for graphical model inference on a suitably
constructed graph.
}
Using Equations~\ref{crf-eq},~\ref{eq-joint}, and~\ref{eq-agree},
we rewrite $\LL(\YA,\vW)$ as
\begin{equation}
\label{eq-simplify}
  \LL(\YA,\vW) = \log \sum_{\vY\in\YA}\exp(\sum_{s,i} \vw_s\vF_s(\vY_{si},\vw_s)) 
     - \sum_{s,i}\log Z(\vX_{si},\vw_s)
\end{equation}
The second part in this equation is the sum of the log partition
function over individual instances which can be computed efficiently
as long as the base models are tractable.  

The first part is equal to the log partition function of a fused
graphical model $G_\aset$ constructed as follows: Initially, each
instance $(s,i)$ creates a graph $G_{si}$ corresponding to its model
$P_s(.)$. For text tasks, this would typically be a chain model with a
node for each token position.  Next, for each clique $C\in\aset$, and
for each pair of triples $(s,i,r), (s',i',r') \in C$, we collapse the
nodes of $r$ in $G_{si}$ with the corresponding nodes of $r'$ in
$G_{s'i'}$.  In Figure~\ref{fig-example}(b) we show an example of such
a fused graph created from the three instances of
Figure~\ref{fig-example}(a) with four cliques in their agreement set.

Let $K$ be the number of nodes in the final fused graph and
$z_1,\ldots,z_K$ denote the node variables. Every node $j$ in the initial graph $G_{si}$
is now mapped to some final node $k\in1,\cdots,K$, and we denote this mapping by
$\map(s,i,j)$.
The log-potential for a component $\vc$ in the fused graph is simply an
aggregate of the log-potentials of the members $\vc'$ that collapsed onto it.
\ignore{
For example, the node log-potential of a final node $z_k$ is:
\begin{equation}
\label{eq-theta1}
\theta_k(z_k) \triangleq \sum_{(s,i,j):\map(s,i,j)=k} \vw_s\vF_s(z_k,\vX_{si},j)  
\end{equation}
and the clique log-potential for a clique $\vc$ is
} 
\begin{equation}
\label{eq-theta2}
\theta_\vc(\vz_\vc) \triangleq  \sum_{(s,i,\vc'):\map(s,i,\vc')=\vc} \vw_s\vF_s(\vz_{\vc'},\vX_{si},\vc')  
\end{equation}
where we extend $\map$ to operate on node-sets as well.
The above $\theta$ parameters now define a distribution over the fused
variables $z_1,\ldots,z_K$ as follows:
\begin{equation}
\label{eq-fusedgraph}
  \prob_\aset(\vz|\vtheta) = \frac{1}{Z_\aset(\theta)}\exp(\sum_\vc \vtheta_\vc(\vz_\vc))
\end{equation}
It is easy to see that the log partition function of this distribution 
is the same as the first term of Equation~\ref{eq-simplify}, 
so we can work with $G_\aset$ instead.
If the set of cliques in $\aset$ is such that the fused graph
$G_\aset$ has a small tree width, we can compute the log partition
function $\log Z_\aset(\theta)$ efficiently. In other cases, we need to
approximate the term in various ways.  We discuss several such
approximations in Section~\ref{sec-approx}.

\subsection{Training algorithm}
The overall training objective of Equation~\ref{eq-collective} is not necessarily
concave in $\vw_s$ because of the agreement term with sums within a
log.  As in ~\cite{LiangKJ07,Ganchev08} it is easy to derive a
variational approximation with extra variables to be solved using an
EM algorithm.  EM will give a local optima if the marginals of the
$\prob_\aset$ distribution can be computed exactly. Since this cannot
be guaranteed for general fused graphs, we also explore the simpler
approach of gradient ascent. In Section~\ref{sec-expt} we show that
gradient ascent achieves better accuracy than EM while being
faster.  The gradient of $\LL(\vY_\aset,\vW)$ is
\begin{equation*}
\nabla\LL(\vY_\aset,\vW) = 
\sum_{s,i,c}\sum_{\vy_c} (\marg_{\aset,\map(s,i,c)}(\vy_c)-\marg_{s,c}(\vy_c|\vX_{si}))\vF_s(\vX_{si},\vy_c,c) 
\end{equation*}
where we use the notation $\marg_{s,c},\marg_{\aset,c'}$ to denote the
marginal probability at $c$ of $\prob_s$ and $c'$ of $\prob_\aset$
respectively. Note that the E-step of EM requires the computation of
the same kind of marginal variables.  These are computed using the
same inference algorithms as used to compute the log partition
function and we discuss the various options next.

\subsection{Approximations}
\label{sec-approx}
We explore two categories of approximations for training when
$\LL(\aset,\vW)$ is intractable. 

\subsubsection{Partitioning the agreement set}
The first category is based on approximating the $\Pr(\YA)$
distribution with product of simpler distributions obtained by
partitioning the set $\aset$.  We partition the agreement set $\aset$
into smaller subsets $\aset_1,\ldots,\aset_R$ such that each
$\prob(\YSp_{\aset_k})$ is easy to compute and
$\cap_k\YSp_{\aset_k}=\YSp_\aset$.  We then approximate
$\Pr(\YA)$ by $\prod_k\prob(\YSp_{\aset_k})$, thus replacing
the corresponding log-likelihood term by $\sum_k \LL(\aset_k,\vW)$.
We explore three such partitionings:
\vskip -0.1in
\paragraph{Clique Agreement}
In this case we have one partition per clique $C \in \aset$.
$G_\aset$ now decomposes into several simpler graphs, where a simple
graph has its nodes fused only at one
clique. Figures~\ref{fig-example}(c)-(f) illustrate this decomposition
for the fused model of Figure~\ref{fig-example}(b).  The probability
$\Pr(\YSp_{\{C\}})$ of agreement on members of a single clique $C$
simplifies to
\begin{equation}
\label{eq-cliqueAdditive}
\Pr(\YSp_{\{C\}}) = \sum_{\vy\in \vY_C} \prod_{(s,i,\vr)\in C} \prob_{s}(\vY_{sir} = \vy|\vX_{si})  
\end{equation}
where $\vY_C$ is set of all possible labelings for any member of $C$,  
and $\prob_s(\vY_{sir} = \vy)$ is the marginal probability of the part $r$ taking the 
labeling $\vy$ under $\prob_s$. 

This approximation is useful for two reasons. First, if the base
models are sequences (e.g.~in typical extraction tasks) and clique
parts $\vr$ are over contiguous positions in the sequence, the fused
graph of $\Pr(\YSp_{\{C\}})$ is always a tree, such as the ones in
Figures~\ref{fig-example}(c)-(f). Second, since for trees we can use
sum-product to compute $\Pr(\YSp_{\{C\}})$ instead of
Equation~\ref{eq-cliqueAdditive}, we can now use arbitrarily long
cliques, instead of choosing unigram cliques which is typically the
norm in extraction applications.
\paragraph{Node Agreement}
We also consider a special case of the clique agreement approximation, called
node agreement, in which each partition corresponds to agreement over
a single variable as in Figure~\ref{fig-example}(e).
\ignore{
For each node $k$ in the fused graph that has been collapsed onto, 
we define a partitioning set $\aset_k$. The probability of agreement, $\Pr(\YSp_{\aset_k})$, 
is calculated as follows:
\begin{equation}
\Pr(\YSp_{\aset_k}) = \sum_{y \in \labelspace} \prod_{\map(s,i,j)=k} \prob_{s}(\vY_{sij}=k|\vX_{si})  
\end{equation}
where $\labelspace$ denotes the label space for a
node. Figure~\ref{fig-example}(e) shows the induced fused model for
agreement on the node `Simpsons'.
}
\paragraph{Instance Pair Agreement}
Another decomposition is based on picking pairs of instances and
defining an agreement set on all cliques which they share.  For the
example in Figure~\ref{fig-example}, graphs marked (g),(h),and (i)
demonstrate the fused graphs arising out of instance pair agreement.
This scheme is expected to be useful when base models exhibit strong
edge potentials. However, unlike for the above two decompositions,
there is no guarantee that the fused graph is a tree (e.g.~graph
(g)). So, approximate inference may be required for some pairs.

\subsubsection{Approximating $Z_\aset(\theta)$}
An alternate way to approximate $\LL(\YA,\vW)$ is to stick with the fused model but approximate the computation
of $Z_\aset(\theta)$. We consider two options:  
\paragraph{Full BP}
In general, any available sum-product inference algorithm like Belief
Propagation and their convergent tree reweighted
versions~\cite{Meltzer09,kolmogorov05Optimality} can be used for
approximating $Z_\aset(\theta)$.  However, these typically require
multiple iterations and can be sometimes slow to converge.
\paragraph{OneStep TRW}
\cite{LiangKJ07} propose a one-step approximation that reduces to a
single step of the Tree reweighted (TRW) family of
algorithms~\cite{kolmogorov05Optimality} where the roles of trees are
played by individual instances. As in all TRW algorithms, this method
guarantees that the log partition value it returns is an upper bound,
but for maximization problems upper bounds are not very useful.

\ignore{
The schemes works by creating a new set of log-potentials from the
$\theta$ log-potentials of Equations~\ref{eq-theta1}
and~\ref{eq-theta2}.

For each instance $(s,i)$ that is part of some clique, we create a new
log-potential $\vthetaC$ as
$\thetaCc\triangleq\frac{\theta_{\map(s,i,c)}}{|\map(s,i,c)|}$ where
$|\map(s,i,c)|$ denotes the number of instances mapped to the same set
in the fused graph as $(s,i,c)$.  It is easy to see that $\theta_c =
\sum_{s,i} \thetaCc$ by construction.  We can now replace $\log Z(\theta)$ with
$\sum_{s,i}\log Z(\thetaC)$.  Note, each $Z(\thetaC)$ is easily
computable since it involves message passing over base tractable
models.
}

A downside of these approaches is that there is no guarantee that the
approximation leads to a valid probability distribution.  For example,
we often observed that the approximate value of $Z_\aset(\theta)$ was
greater than $\sum_{s,i}\log Z(\vX_{s,i},\vw_s)$ causing the
probability of agreement to be greater than 1.

To summarize, we would ideally like to optimize the agreement-based objective in 
Equation~\ref{eq-collective} exactly by working with the equivalent fused graphical 
model of Equation~\ref{eq-fusedgraph}. Due to intractability, we discussed various 
ways to decompose the agreement term or approximate the corresponding fused model. 
As we shall show in Section~\ref{sec-expt}, 
when there are noisy cliques in the agreement set, the tractable decompositions turn out to be
much more robust than 
methods that approximate the fused model created from erroneous cliques. 

\input{cliques}

\input{related}

\input{expt}

\input{concl}

\bibliographystyle{plain}


\end{document}

%% file: abstract.tex
\begin{abstract} 
We consider the problem of jointly training structured models for
extraction from sources whose instances enjoy partial overlap.  This
has important applications like user-driven ad-hoc information
extraction on the web. 
Such applications present new challenges in terms of the number of
sources and their arbitrary pattern of overlap not seen by earlier
collective training schemes applied on two sources.
We present an agreement-based learning framework and alternatives
within it to trade-off tractability, robustness to noise, and extent
of agreement.  We provide a principled scheme to discover low-noise
agreement sets in unlabeled data across the sources.  Through
extensive experiments over 58 real datasets, we establish that our
method of additively rewarding agreement over maximal segments of text
provides the best trade-offs,
and also scores over alternatives such as collective inference, staged
training, and multi-view learning.

\ignore{
We present the problem of jointly training extraction models 
for multiple sources whose instances enjoy partial content overlap.
This has important applications like user-driven ad-hoc
information extraction on the web. We employ agreement-based learning 
that biases the models to agree on the labelings of overlapping 
regions. We present various ways to decompose the intractable training objective 
into simpler terms that also guard against any noise in agreement. We also 
provide a principled scheme to derive agreement sets from partial overlaps
in $>$2 sources. Last, through extensive experiments over 58 real datasets, we 
establish that agreement-based learning along with our approximations and 
agreement-set generation scheme are superior to other known alternatives like 
collective inference, and multi-view learning.
}

\end{abstract} 

%% file: intro.tex
\label{sec-intro}
This paper addresses the problem of training multiple structured
prediction models that share an output space but differ in their input
data and feature space. Further, labeled data in each source is
limited, but unlabeled data over the different sources overlap
partially.
This scenario is applicable in many text modeling tasks such as
information extraction, dependency parsing, and word alignment.  These
tasks are increasingly being deployed in settings where supervision is
limited but redundancy is abundant. A concrete motivation for our work
comes from recent efforts to support rich forms of structured
query-answering on the Web~\cite{carlson10Coupled,Cafarella08Web}. A
typical subtask here is building extraction models over multiple Web
documents starting from a small seed of user-provided structured records.

Recently, many learning paradigms have been proposed to exploit the
relatedness of multiple models. On one end of the spectrum we have
collective
inference~\cite{sutton04skip,bunescu04,krishnan06:effective,finkel05:Incorporating,gupta07abbr}
where each model is trained independently but prediction happens
jointly to encourage agreement on overlapping content. On the other
end are methods like multi-view learning~\cite{Ganchev08,ganchev09}
and agreement-based learning~\cite{LiangKJ07,LiangTK06} that formulate
a single objective to jointly train all models. Then there are methods
in-between that train models sequentially or
alternately~\cite{BrefeldBS05,carlson10Coupled}.  Our problem is
different from traditional multi-view learning where multiple models
are trained on different views of a {\em single} data source. However, by
treating the different contexts in which each shared portion resides
as a different view, we can apply multi-view learning to
this problem.  We elaborate on this and other alternatives in
Section~\ref{sec-related}.

In agreement-based learning~\cite{LiangKJ07,LiangTK06} the goal is to
train multiple models so as to maximize the likelihood of the labels
agreeing on shared variables.  However, these assume that all models
need to agree on the same set of variables --- this trivially holds
for two sources where these methods have been applied.  In our
application the number of sources is often as large as 20.  As number
of sources increase, there is a bewildering number of ways in which
they overlap.  This makes it challenging to devise objectives that
maximally exploit the overlap while accounting for noise in the
agreement set and intractability of training.  We are aware of no
study where such issues are addressed in the context of jointly
training more than two sources with partial overlap.

In this paper we propose an agreement-based model for training
multiple structured models with arbitrary partial overlap among the
sources.  We propose several alternatives for enforcing agreement
ranging from singleton variables, to groups of contiguous variables,
to global models that lead to giant agreement graphs.  For the task of
information extraction, we present a strategy for selecting the unit
of agreement that leads to a significant reduction in the noise in the
agreement set compared to the existing na\"ive approach for choosing
agreement sets.\ignore{picking any repeated non-stop word as an
agreement set.}  We present an extensive evaluation on 58 real-life
collective extraction tasks covering a rich spectrum of data
characteristics.  This study reveals that agreement objectives that
are additive over smaller components provide the best accuracy because
of robustness against noise in the agreement term, while providing a
tractable inference objective.
\ignore{
We show
that our method of generating clique agreement terms reduces noise
from 17.3\% to 5.6\%. 
}
\ignore{
We provide a concrete motivation from building collective information
extraction models during structured query answering on the
Web. Suppose a user is interested in compiling list of structured
records, say a list of award winning children's books with fields like
author, book title, year, and name of award. The user starts
with only a few sample records and finds several HTML list
sources that provide more records.  However, each HTML source is
unstructured and an extraction model such as a CRF is used to convert
it into structured records. This gives rise to a collective learning
problem with the following characteristics:
Each source has a small number of labeled records from the starting
examples, there is significant overlap in the content of various
sources, but each source has its own distinct style making it
necessary to train separate extraction models for each source.
}

\ignore{
Introduce the problem --- separate structured models over separate corpora. 
But the separate corpora have partial content overlap, largely over their unlabeled data. 
Example here. Since the corpora are separate this is not multi-task learning, and since 
overlap is not complete, it is not traditional multi-view either, although 
some ideas might apply. We expect
}

%% file: cliques.tex
\section{Generating the agreement set}
\label{sec-agree}

In this section we discuss our unsupervised strategy for finding agreement sets. But first we stress that 
the importance of this step cannot be overstated. As we show in Section~\ref{sec-expt}, 
even the best collective training schemes are only as good as their agreement set.
This has interesting parallels with other learning tasks e.g.~semi-supervised learning,
where recent work has shown the importance of creating good neighborhood graphs~\cite{jebara09Graph}. 

Traditional collective extraction methods
have not focused on the process of finding quality agreement sets.
These methods usually form a clique from arbitrary repetitions of unigrams
~\cite{sutton04skip,finkel05:Incorporating,krishnan06:effective}.
This is inadequate because of two reasons. First, any strong 
first order dependencies cannot be transferred with only unigram cliques. 
Second, blindly marking repetitions of a token/n-gram as a clique can
inject a lot of noise in the agreement set. 

Instead we use a more principled strategy. We make the working assumption that significant content overlap among 
sources is caused by (approximate-)duplication of instances. So we assume that each instance has a hidden 
variable with value equal to its `canonical instance value'. Instances inside a source will have different values of 
this variable (as duplicates are rare inside a source), whereas these values will be shared across sources, thus forming 
clusters. Assume for now that these clusters are known. Given such a cluster of deemed duplicates, 
we find maximally long segments that repeat among the instances in the cluster, and add
one clique per such segment to the agreement set.  Segment repetitions outside the cluster are considered as false matches
and ignored.
\ignore{All these operations can be optimized using standard reverse index primitives.}

The task of optimally computing the clusters essentially reduces to the NP-hard multi-partite 
matching problem with suitably defined edge-weights. 
We tackle this by employing the following staged scheme: First, we order the sources 
using a natural criteria such as average pairwise similarity with the other sources. Each instance in the first source forms a 
singleton cluster. In stage $s$, we find a bipartite matching between source $s+1$ and the clusters formed by the 
first $s$ sources. An instance $i$ in source $s+1$ will be assigned to the cluster to which it is matched. 
Unmatched instances form new singleton clusters. The edge-weight between an instance $i$ and a cluster is defined as 
the best similarity of $i$ with any member instance of the cluster.

When our assumption of instance duplication does not hold, say when each instance is an arbitrary natural language sentence, 
the bipartite matching scores will be low and we revert to the conventional clique generation scheme. 
As we shall see in Section~\ref{sec-expt}, our strategy generates much better agreement cliques in practice.

%% file: related.tex
\section{Relationship with other approaches}
\label{sec-related}
We now review various approaches relevant to collective training with partially overlapping sources. We
omit Agreement-based learning as it has already been discussed in Sections~\ref{sec-intro} and \ref{sec-approx}.

\subsection{\label{sec-PR}Posterior regularization (PR)}
The PR framework~\cite{ganchev09TR} trains a model with task-specific linear constraints on the posterior.
The constrained optimization problem is solved via the EM algorithm on its variational form.
PR has been shown to have interesting relationships with similar frameworks~\cite{Liang08Learning,mann08Generalized,chang07Guiding}.

The aspect of PR most relevant to us is its application to multi-view learning~\cite{Ganchev08}. Then the PR constraints 
translate to minimizing the Bhattacharayya distance between the various posteriors. This has two key differences with our setting. 
First their agreement set is at the level of full instances instead of arbitrary sub-parts. Moreover, their agreement set has no noise 
because the instances across views are known duplicates instead of assumed ones like in ours. The second and more interesting difference 
is that of the agreement term.

\begin{figure}
\label{fig-loss}
\begin{center}
\subfigure{\includegraphics[width=0.40\textwidth]{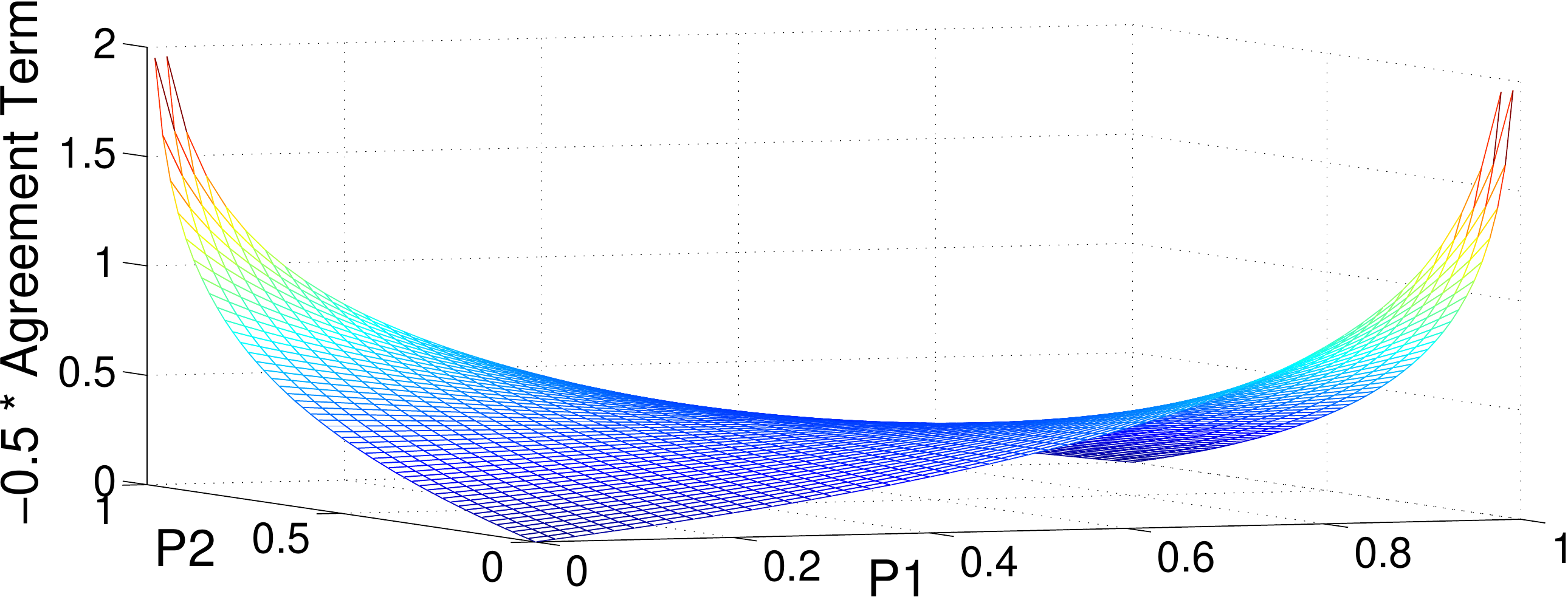}}
\subfigure{\includegraphics[width=0.40\textwidth]{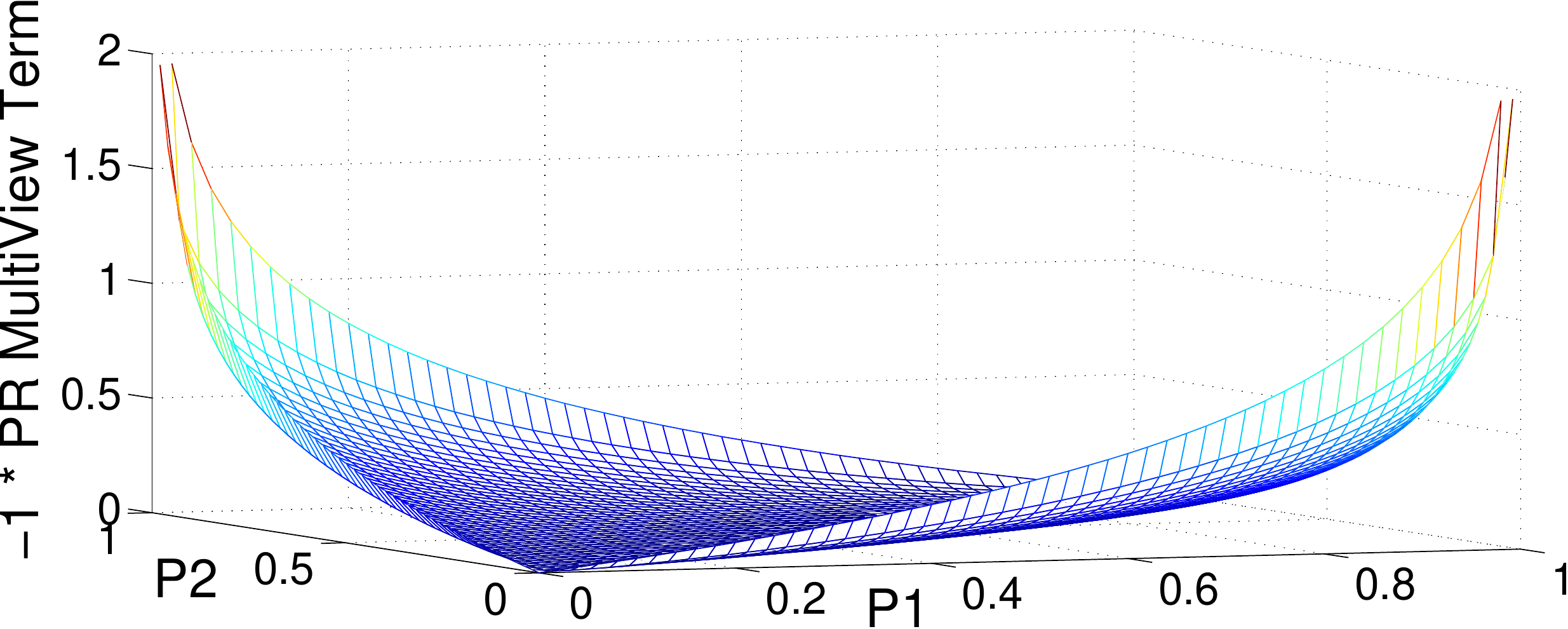}}
\caption{Comparison of the agreement (left) and multi-view (right)
losses over two binomial posteriors}
\end{center}
\vskip -0.2in
\end{figure}

Assuming that we have only two sources $s$ and $s'$, with only one shared clique $\vc$, training the two models is the same 
as learning in the presence of
two-views of $\vc$. The agreement term under PR would be 
$\log \sum_{\vy_\vc} \sqrt{P_s(\vy_\vc)P_{s'}(\vy_\vc)}$, where $P_.(\vy_\vc)$ is the marginal of $\vc$.
This is maximized when the two marginals are identical. In contrast, 
our agreement term of $\log \sum_{\vy_\vc} P_s(\vy_\vc)P_{s'}(\vy_\vc)$ is maximized when the marginals are  identical
{\em and peaked}. 
If both the base models are strong, their marginals will be 
almost peaked, resulting in little difference between the two terms. 
But a difference arises in the asymmetric case when one model is
strong and peaked and the other is weak and flat. One possible maxima
for the two-view term would be the strong model flattening out and
becoming identical to the weaker one.  As our agreement term is
averse to flat marginals, it will avoid this maxima. 
Figure~\ref{fig-loss} illustrates the difference between the
two terms for two binomial distributions.

Given such a relationship between the two terms, we compare
the performance of our algorithms with the multi-view algorithm in
Section~\ref{sec-expt}. The multi-view objective will be
optimized via EM because the gradient cannot be computed
tractably for structured models.


\subsection{\label{sec-stage}Label transfer}
Another set of approaches deal with transferring
labeled data from one source to the other. One such inexpensive
approach is an asymmetric staged strategy of training the model for a
more confident source first, transferring its certain labels to the
next source and so on. This requires a good ordering of
sources to control error-cascades.  In Section~\ref{sec-expt}, we show
that even with suitable heuristics, this scheme suffers from huge
performance deviations.  Similar label transfer ideas have been employed in 
training rule-based extraction wrappers ~\cite{carlson10Coupled}.

More sophisticated methods in this class include
CoBoosting~\cite{collins99Unsupervised}, Co-Training~\cite{blum99Combining},
and the two-view Perceptron~\cite{BrefeldBS05} that train two models
in tandem by each model providing labeled data for the other. 
A detailed comparison of these models in \cite{ganchev09TR} show that
these methods are less robust than methods that jointly train all
models.

\subsection{\label{sec-ci}Inference-only approaches}
Another option is to only train the base models, and perform any
corrections at runtime through collective inference. Such strategies
have been used on a variety of NLP tasks
~\cite{sutton04skip,finkel05:Incorporating,krishnan06:effective}. 
These methods usually end up using cliques only over
unigrams, with little focus on controlling their noise.
The most common practice is marking
arbitrary repetitions of a token as a clique. As we show in
Section~\ref{sec-expt}, our collective training algorithms are
significantly better than collective inference, even with identical
agreement sets.  A prime limitation of inference-only approaches is
that they cannot transfer the benefits of overlap to other instances
which do not overlap.

%% file: expt.tex
\section{Experimental evaluation}
\label{sec-expt}
We present extensive experiments over several real datasets
covering a rich diversity of data characteristics. Our first
set of experiments seek to justify collective training by showing
substantial benefits over base models, and alternatives like staged
training and collective inference discussed in Sections~\ref{sec-stage}
and \ref{sec-ci} respectively.  Second, we study our collective
training approach in detail by comparing the accuracies of the various
approximations made in Section~\ref{sec-approx}. Third, we demonstrate
the importance of choosing high quality agreement sets by comparing
various set-generation schemes. Finally, we make a case that our
simple gradient ascent algorithm is as accurate as existing
traditional EM-based approaches~\cite{LiangKJ07,Ganchev08} while being
considerably faster.

\noindent {\textbf{Datasets: }} We use a corpus of 58 real datasets,
 each comprising multiple HTML
lists. All lists in a dataset contain semi-structured instances
relevant to a dataset-specific relation e.g.~University mottos,
Caldecott medal winners, movies by James Cagney, Supreme court cases
etc.
\ignore{
Figure~\ref{fig-datasetExample} shows fragments of two lists from the
Caldecott dataset.  Columns of the relation define a label set for the
dataset, e.g.~Year, Illustrator and Book for the Caldecott dataset.  }
The 58 datasets exhibit a wide spectrum of behavior in terms of their
base accuracy, number of sources, number of cliques per instance,
their noise levels, and so on.  For ease of presentation, we partition
these 58 sources into ten groups by a paired criteria --- base
accuracy and relative size of the agreement sets. We create five bins
for base accuracy values: 50--60, 60--70, and so on, and two bins for
agreement set: ``M'' (many) when there are more than 0.5 cliques per
instance and ``F'' (few) otherwise.  Table~\ref{tab-data} lists
for each of the ten groups the number of datasets (\#), average
 number of sources ($S$), number of labels
($|\labelspace|$), number of cliques ($|\aset|$), instances, base F1
  score, and noise in the agreement set $\aset$.  The last row in
  the table that lists the standard deviation of these values over all
  58 sources illustrates the diversity of the dataset.

\paragraph{Task}
For each dataset, we mimic a user query by seeding with a handful of
structured records.  These are used to generate labeled data out of
matching instances in each list of the dataset. The goal is to learn a
robust model for each list and extract more instances from it.  All
comparisons are with 3 and 7 seed records only. Bigger training sets
are not practical in this task as the seed structured records are 
provided through a manual query.
All our numbers are averaged over five random selections of
the seed training set.  Our base model is a conditional random field
trained using standard context features over the neighborhood
of a word, along with class prior and edge features.
Our ground truth consists of every token manually labeled with
a relevant dataset-specific label. Using this ground truth, we denote a 
clique as pure if all its members agree on their true labels, and noisy otherwise.
We measure model accuracy by the F1
score of the extracted entities. We set $\lambda$ using a validation
set.

\begin{table}
\begin{center}
\begin{small}
\begin{tabular}{lrrrrrrr}
\hline
&\# Datasets &S& $|\labelspace|$ &$|\aset|$&Instances& F1 Base& $|\aset|$ Noise \\
\hline
50F &2&9&4.0&23&75   &55.23&0.10\\
50M&3&11&4.3&202&223 &54.59&0.11\\
60F &4&6&3.5&147&409 &67.53&0.07\\
60M&4&14&4.5&235&344 &67.55&0.12\\
70F &3&9&5.3&146&346 &76.67&0.35\\
70M&14&10&4.4&413&336&75.17&0.21\\
80F &9&14&4.0&172&575&85.75&0.04\\
80M&7&13&4.0&959&831 &85.9&0.11\\
90F &6&10&4.0&154&440&94.95&0.04\\
90M&6&15&4.0&436&493 &95.71&0.13\\\hline
All&58&11&4.2&348&451&79.56&0.15\\
Std&0&5.8&1.1&500&432&12.24&0.14\\
\hline
\end{tabular}
\end{small}
\end{center}
\caption{\label{tab-data}Properties of the datasets}
\end{table}

\ignore{ We note that our focus being collective training, we assume
  that the process of generating the labeled data from user input is
  error-free. }

\subsection{Benefit of collective training}
We first compare collective training, collective inference, and staged
label-transfer methods with the base model, starting with only
three labeled instances from the user. We chose the Clique agreement method (\cliqueA)
for collective training, and used the same agreement set for the
collective training and collective inference. 
\begin{figure}
\begin{center}
\includegraphics[width=0.45\textwidth]{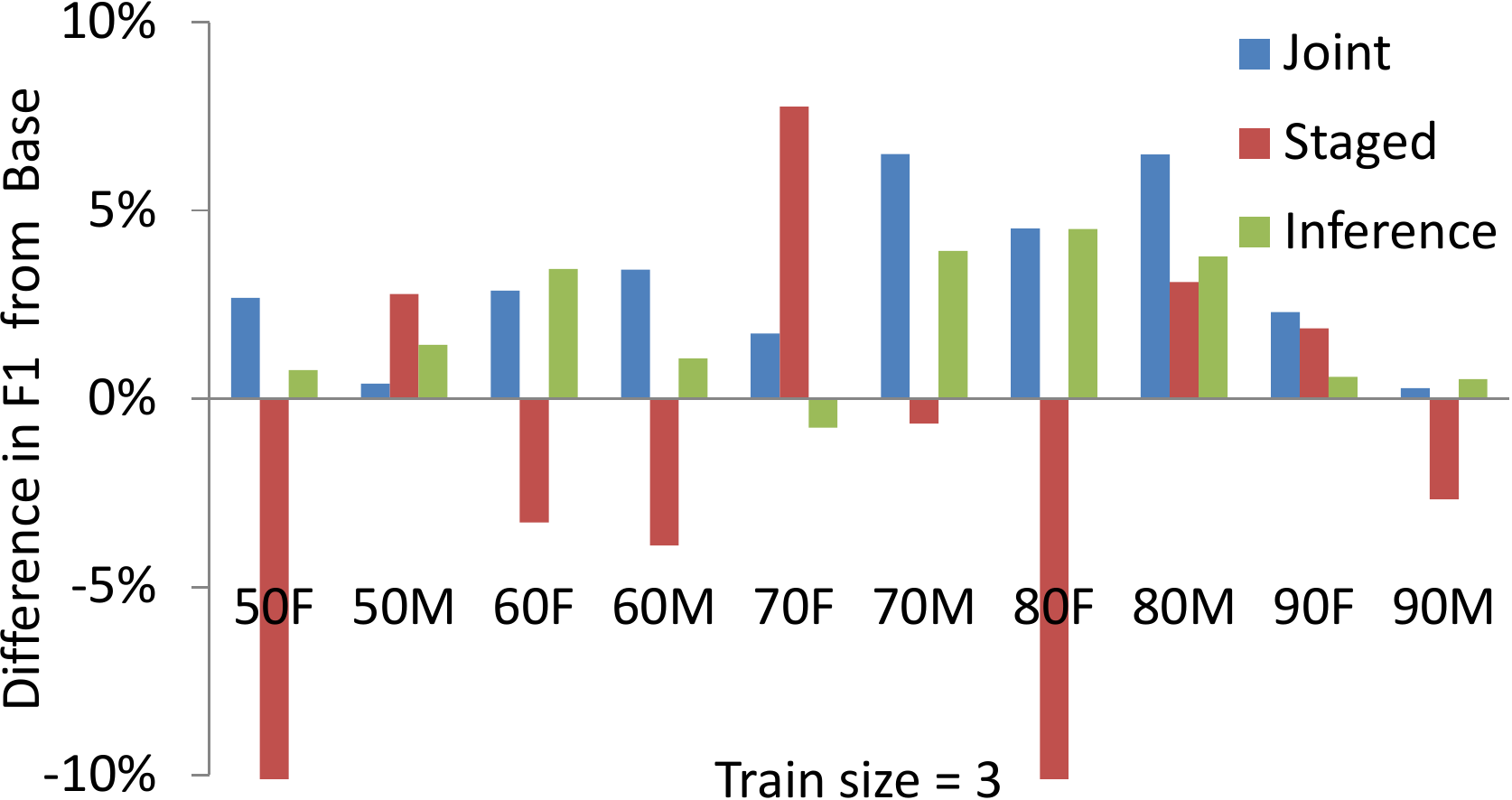}
\caption{Comparing Clique Agreement against the Collective Inference and Staged approaches}
\label{fig-expt1}
\end{center}
\vskip -0.2in
\end{figure}

Figure~\ref{fig-expt1} shows the gains and losses of the three
methods over the base model for each of ten groups.  Collective
training clearly performs the best, and its gains are specially large
for datasets whose base accuracy is in the 60--80\% range, and which
have big agreement sets.  Overall, its F1 is 87.9\% in
contrast to the base accuracy of 83.7\%.  Even with a training size of
seven, F1 improves from 87.4 to 89.2 (not shown in the figure).  In
contrast, the staged approach overall performs worse than Base and
shows large swings in accuracy across datasets.  It is highly
sensitive to the ordering of sources, and 
the hard label-transfer often causes error-propagation to all downstream
sources.  Collective inference improves accuracy in a few cases but
overall provides only a small gain of 0.3\% beyond Base.  

\subsection{Comparing collective training objectives}
We now compare the various approximations of the agreement term ---
\cliqueA, \nodeA, \pairA, \fullBP\ (Full BP), and \trwOne\ (OneStep TRW) as described in
Section~\ref{sec-approx}.    Table~\ref{tab-expt}
shows the gains in F1 for all the approaches over the base model.

\begin{table}
\begin{center}
\begin{small}
\begin{tabular}{lr|rrrrr|r}
\hline
&    &\multicolumn{5}{|c|}{Agreement} & \dist  \\
Data&Base&\cliqueAS&\nodeAS&\pairAS&\fullBP&\trwOne & EM\\
\hline
\multicolumn{7}{c}{Train size = 3}\\
{\bf All}& 83.3 & {\bf 4.2} & 3.9 & 2.6 & 2.6 & 2.1 & 3.7\\
\hline
50F & 55.2 & 2.7 & {\bf 3.5} & 2.9 & 2.9 & {\bf 3.5} & 1.0\\
50M & 54.6 & 0.6 & 0.9 & 1.3 & 1.1 & {\bf 4.5} & 3.6\\
60F & 66.9 & {\bf 2.9} & 2.6 & 0.8 & 0.6 & 1.5 & 1.5\\
60M & 67.3 & {\bf 3.4} & 2.3 & 1.8 & 2.2 & -0.1 & 3.4\\
70F & 73.5 & {\bf 1.7} & 1.2 & 1.4 & 1.0 & 0.7 & 1.1\\
70M & 76.1 & 6.5 & 5.8 & 3.8 & 4.5 & 3.7 & {\bf 6.9}\\
80F & 85.6 & {\bf 4.5} & 4.1 & 3.7 & 3.5 & 0.2 & 4.4\\
80M & 86.6 & {\bf 6.5} & 6.0 & 3.8 & 3.4 & 3.6 & 4.5\\
90F & 94.3 & {\bf 2.3} & 2.1 & 0.5 & 1.1 & 1.2 & 1.7\\
90M & 96.1 & 0.3 & {\bf 0.6} & -0.1 & 0.0 & {\bf 0.6} & 0.4\\
\hline
\multicolumn{7}{c}{Train size = 7}\\
{\bf All}& 87.3 & 1.8 & {\bf 2.0} & 1.0 & 0.9 & 0.7 & 2.1\\
\hline
50F & 52.5 & 4.2 & {\bf 5.6} & 4.8 & 4.8 & 4.2 & 3.4\\
50M & 63.4 & 0.0 & 0.1 & 0.9 & 0.4 & {\bf 3.5} & 2.7\\
60F & 76.2 & {\bf 1.9} & 1.5 & 0.1 & 0.1 & 0.5 & 1.2\\
60M & 75.0 & 1.3 & {\bf 2.6} & 0.9 & 0.3 & -1.5 & 2.4\\
70F & 79.6 & 2.3 & {\bf 3.3} & 0.1 & 0.0 & -0.8 & 2.5\\
70M & 82.4 & 3.2 & 3.5 & 2.2 & 2.2 & 1.7 & {\bf 3.8}\\
80F & 90.3 & {\bf 1.5} & 1.4 & 1.1 & 1.2 & 0.6 & 1.6\\
80M & 90.5 & {\bf 2.7} & 2.6 & 1.3 & 1.0 & 0.9 & 2.4\\
90F & 96.6 & 0.6 & 0.2 & 0.1 & 0.2 & 0.2 & {\bf 0.8}\\
90M & 96.5 & 0.3 & {\bf 0.7} & 0.2 & 0.2 & -0.2 & 0.5\\
\hline
\end{tabular}
\end{small}
\caption{\label{tab-expt}Comparing different training approximations
  in terms of F1 accuracy gain over the base model.}
\end{center}
\end{table}

Observe that \cliqueA\ and \nodeA\ agreement are two of the best
performing methods. We explore two possible reasons for why they score
over other approaches that fuse the influence of multiple cliques.
One partial explanation is that 15\% of the cliques in our agreement set are noisy. 
In such a case, fused methods would try hard at
maximizing the likelihood of a wrongly fused graph. In contrast, the
\cliqueA\ and \nodeA\ agreement models decompose over cliques, so they
can choose to ignore the terms corresponding to erroneous cliques
during optimization. A second reason common to all the losing
approaches is the inexact nature of the optimization of their training
objectives.  To understand which of these is a plausible reason, we
remove all noisy cliques using the ground truth and compare
\cliqueA\ and \pairA\ agreement.  Accuracy improves by less than 0.6
F1 in both, and \cliqueA\ continues to score over \pairA.  This
indicates that inexact gradient computation is perhaps a major reason
why more complex fused approaches perform worse.

\begin{figure}
\begin{center}
\subfigure[Difference in \cliqueAS\ and \nodeA\ against noise]{\label{fig-additiveScatter}\includegraphics[width=0.3\textwidth]{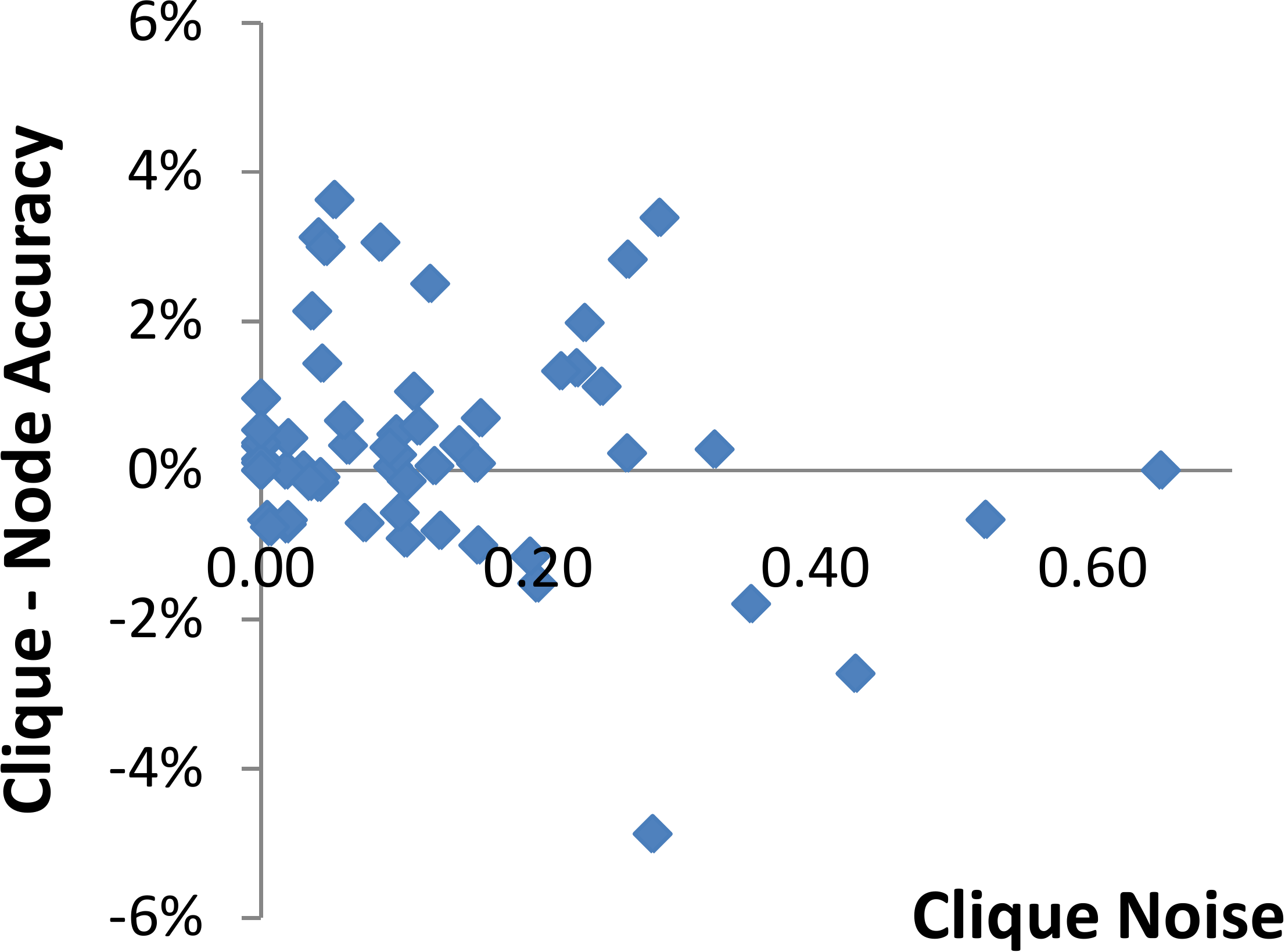}}
\subfigure[Accuracy of \nodeA\ with clique generation methods of varying noise]{\label{fig-noise}\includegraphics[width=0.3\textwidth]{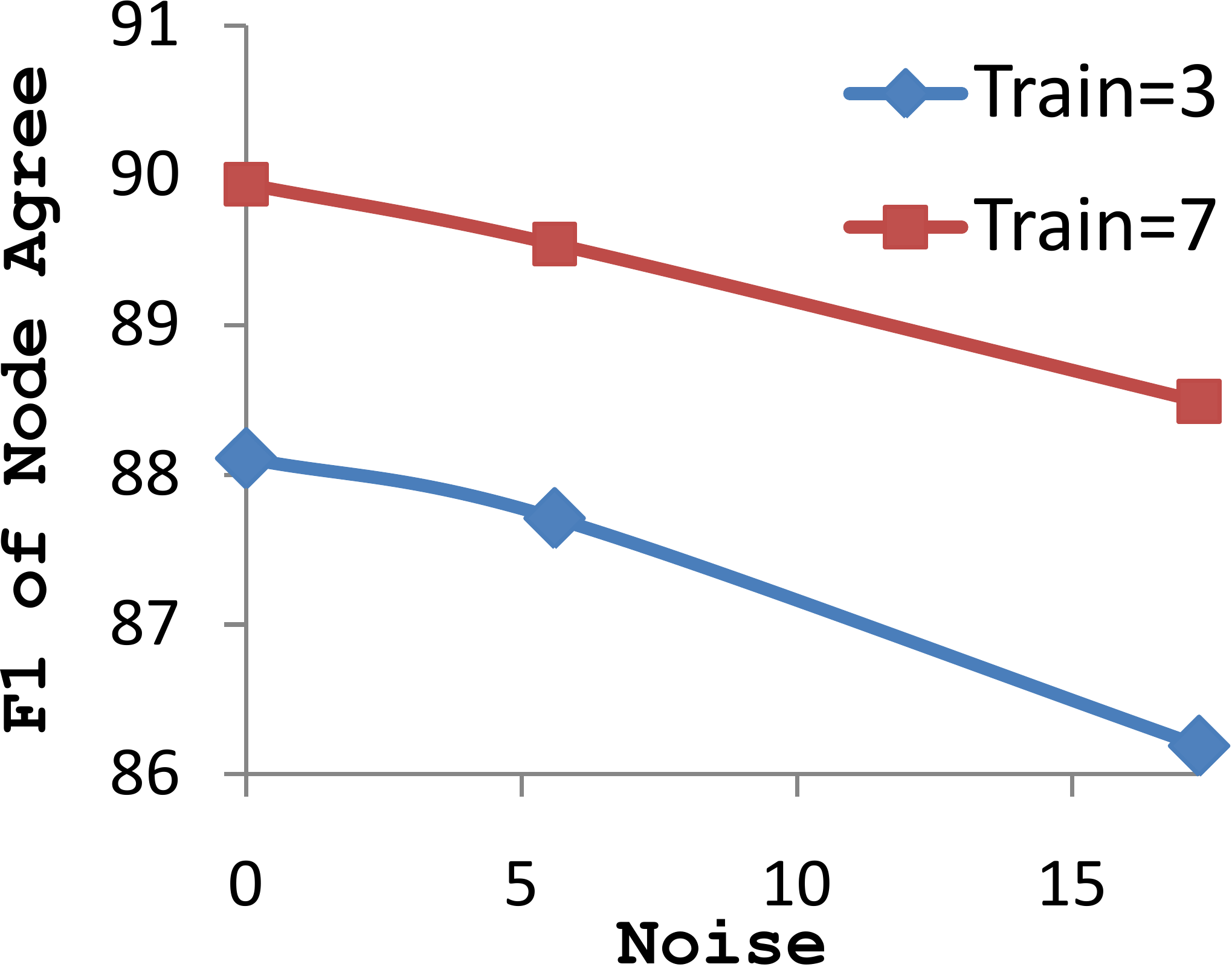}}
\caption{Effect of clique noise on collective training}
\end{center}
\vskip -0.2in
\end{figure}
We also see that there is little difference between the \cliqueA\ and
\nodeA\ agreement models. While one possible reason is the weakness of
any first-order dependency in the true model, we find another
interesting reason for this behavior. We note that in a general n-gram
agreement clique, only a few positions might be erroneous. For
example, the 15\% noise measured at segment level reduces to 5.6\% at
position level.  Since \nodeA\ decomposes the clique over positions, it
can ignore wrong positions during optimization and be more robust
against noisy cliques. We corroborate this in
Figure~\ref{fig-additiveScatter}, where for each of the 58 datasets,
we plot the difference in F1 of \cliqueA\ and \nodeA\ versus clique
noise in the dataset's agreement set. We observe that whenever
\cliqueA\ performs sufficiently worse than \nodeA\ there is high noise
in the cliques. In low noise settings, \cliqueA\ is often much better
than \nodeA.

\subsection{Noise in the Agreement Set}
As discussed in Section~\ref{sec-agree}, it is important to choose
high quality agreement sets. Figure~\ref{fig-noise} shows the F1
scores of \nodeA\ under three clique generation schemes of varying
noise.  The rightmost points are for the conventional practice of
choosing arbitrary unigram repetitions as cliques and has a noise of
17.3\% at position level. The middle point is our method of clique generation where we
reduce the noise to 5.6\% and the leftmost are ideal cliques with zero
noise obtained by using the ground truth to remove all noisy unigrams.
We find that our clique selection method enjoys accuracy very
close to that with noise-free cliques and the accuracy
with carelessly chosen cliques is much lower.

\subsection{Comparison with EM-based approaches}
In Section~\ref{sec-PR} we described how the PR framework
\cite{Ganchev08} is applicable to our problem. We
show its results in the last column of Table~\ref{tab-expt}. The
accuracy of PR is comparable to the \cliqueA\ method
showing that distance-based and likelihood terms serve similar goals
in our setting.  However, the \dist\ approach is more than four times
slower than our likelihood objective maximized using gradient ascent.
The \dist\ objective requires the EM algorithm for training. In
typical feature-based structured models, the M-step tends to be 
expensive and it is best not wasted on working with fixed E-values.
To evaluate the tradeoffs between EM and gradient-based training we
also ran the EM algorithm of \cite{LiangKJ07} whose gradient-based
version we call \trwOne\ in Table~\ref{tab-expt}.  We found the EM
trainer (not shown) to have an F1 0.4\% less than \trwOne\ and also a
factor of two slower.

%% file: concl.tex
\section{Conclusion}
We presented a framework for jointly training multiple extraction models exploiting
partial content overlap across sources. Partial overlap opens up a slew of problems --- choosing 
a noise-free agreement set, a training objective or its approximation, and an optimization 
algorithm. We showed that while decomposing the agreement term over cliques provides a tractable yet 
accurate method of agreement, it also turns out to be more robust against clique noise than methods 
that approximate the fused graph. We also presented a strategy for computing clean agreement sets that is far superior 
to the na\"ive alternative. Through extensive experiments on various real datasets we showed that our agreement-term 
decompositions on cliques and positions are more robust, accurate, and faster than alternatives like multi-view learning.